# TOWARDS CULTURAL BRIDGE BY BAHNARIC-VIETNAMESE TRANSLATION USING TRANSFER LEARNING OF SEQUENCE-TO-SEQUENCE PRE-TRAINING LANGUAGE MODEL

Phan Tran Minh Dat[1,3], Vo Hoang Nhat Khang[1,2,3], Quan Thanh Tho[1,2,3,*]

[1] Faculty of Computer Science and Engineering, Ho Chi Minh City University of Technology (HCMUT), 268 Ly Thuong Kiet Street, District 10, Ho Chi Minh City, Vietnam

[2] Office for International Study Programs, Ho Chi Minh City University of Technology (HCMUT), 268 Ly Thuong Kiet Street, District 10, Ho Chi Minh City, Vietnam

[3] Vietnam National University Ho Chi Minh City, Linh Trung Ward, Thu Duc District, Ho Chi Minh City, Vietnam

[*] Corresponding author: qttho@hcmut.edu.vn

## Abstract

This work explores the journey towards achieving Bahnaric-Vietnamese translation for the sake of culturally bridging the two ethnic groups in Vietnam. However, translating from Bahnaric to Vietnamese also encounters some difficulties. The most prominent challenge is the lack of available original Bahnaric resources source language, including vocabulary, grammar, dialogue patterns and bilingual corpus, which hinders the data collection process for training. To address this, we leverage a transfer learning approach using sequence-to-sequence pre-training language model. First of all, we leverage a pre-trained Vietnamese language model to capture the characteristics of this language. Especially, to further serve the purpose of machine translation, we aim for a sequence-to-sequence model, not encoder-only like BERT or decoder-only like GPT. Taking advantage of significant similarity between the two languages, we continue training the model with the currently limited bilingual resources of Vietnamese-Bahnaric text to perform the transfer learning from language model to machine translation. Thus, this approach can help to handle the problem of imbalanced resources between two languages, while also optimizing the training and computational processes. Additionally, we also enhanced the datasets using data augmentation to generate additional resources and defined some heuristic methods to help the translation more precise. Our approach has been validated to be highly effective for the Bahnaric-Vietnamese translation model, contributing to the expansion and preservation of languages, and facilitating better mutual understanding between the two ethnic people.

## 1. Introduction

The Bahnar people, also known as *Ba-na* (pronounced [ɓaː˧naː˧] in Vietnamese), represent a distinct ethnic minority within Vietnam's diverse ethnic mosaic. Currently, a significant part of efforts to integrate the identity into mainstream society involves translating Bahnaric documents to Vietnamese, with contributions from both government authorities and local communities. Thanks to remarkable advances in artificial intelligence in general and Neural Machine Translation (NMT) in particular, the translation process has become much more accurate and fluent.

Machine translation systems based on deep learning models have achieved milestones with the emergence of techniques such as Word2Vec (Mikolov et al. 2013) [1], Convolutional Neural Networks (CNNs), Recurrent Neural Networks (RNNs) or Long Short-Term Memory (LSTM) (Hochreiter and Schmidhuber 1997) [2]. Later, the RNN search architecture introduces an approach to implementing variable-length representations - the attension mechanism. Seq2Seq [3] pioneered the potential of deep NMT architecture. Models like ByteNet (Kalchbrenner et al. 2016) [4], ConvSeq2Seq (Gehring et al. 2017) [5], and Transformer also utilize multi-layered networks.

Data augmentation (DA) was first widely applied in the computer vision field and then used in natural language processing (NLP), achieving improvements in many tasks. Many DA strategies for NLP, from rule-based manipulations (Zhang, Zhao, and LeCun 2015) [6] to more complex generative systems (Liu et al. 2020) [7], have been developed despite challenges associated with text.

However, translating Bahnaric to Vietnamese is a challenging task due to the extreme scarcity of resources for the Bahnaric language, including the lack of available language corpora and limited vocabulary. Robinson et al. conducted a decision tree analysis of language features and their correlation with the effectiveness of LLMs in machine translation, and found out that even ChatGPT is particularly disadvantaged when translating low-resource languages (Robinson et al. 2023) [8].

To address these challenges, we propose using transfer learning principles with carefully selected pre-trained models. This approach aims to enhance translation quality and optimize computational efficiency. In order to tackle the resource scarcity issues, we incorporate specialized augmentation strategies into the translation task, fine-tune our model for optimal performance in translating between these two languages, as long as developing a multiple heurisitc approaches to enhance the translation.

In this study, our contributions are to present a Bahnaric-Vietnamese language translation which utilizes the strengths of modern NMT based on the pre-trained BARTpho architecture and the linguistic similarities between the two languages, as long as introducing appropriate data augmentation methods for the Bahnaric language and some heuristic methods to improve the translation process.

# 2. Methods and Implementation

## 2.1. Overview of Pipeline

Figure 1 illustrates the complete process of our Bahnaric-Vietnamese language translation system. Bahnaric is considered a low-resource language, indicating a scarcity of linguistic data and resources for natural language processing tasks. Consequently, traditional machine translation methods that depend on extensive parallel corpora for training are neither feasible nor effective for this task. Nevertheless, Vietnamese and Bahnaric share several grammatical features, such as word order, morphology, and syntax, which can be leveraged to enhance the machine translation process and improve output quality. Our proposed method employs a chunking translation approach, combining word mapping with the fine-tuning of a pretrained language model for machine translation. The proposed method features an end-to-end pipeline consisting of two main phases:

- **Segmentation phase:** A Bahnaric sentence is passed as an input and is splitted into meaningful words using a word segmenter which we previously built based on the characteristics of the language. These words are classified into anchors or chunks, each of which will be further processed in different ways in the next stage. Anchors are words or phrases which can be directly mapped to Bahnaric language using Bahnaric-Vietnamese dictionary, while chunks needs further translation. The output of this phase is a list of the anchors and chunks which is identified from the input sentence.
- **Mapping phase:** The list of anchors and chunks identified from the previous phase is employed different techniques to map all of them to Vietnamese. For anchors, the words which exist in the Bahnaric-Vietnamese dictionary, they are converted to the corresponding Vietnamese meaning. For chunks, we developed a fine-tuned BARTpho model (a sequence-to-sequence model for Vietnamese) (Tran, Le, and Nguyen 2022) [] to translate them into Vietnamese language phrases. The output of this phase is a list of translated segments from the the original list of anchors and chunks.

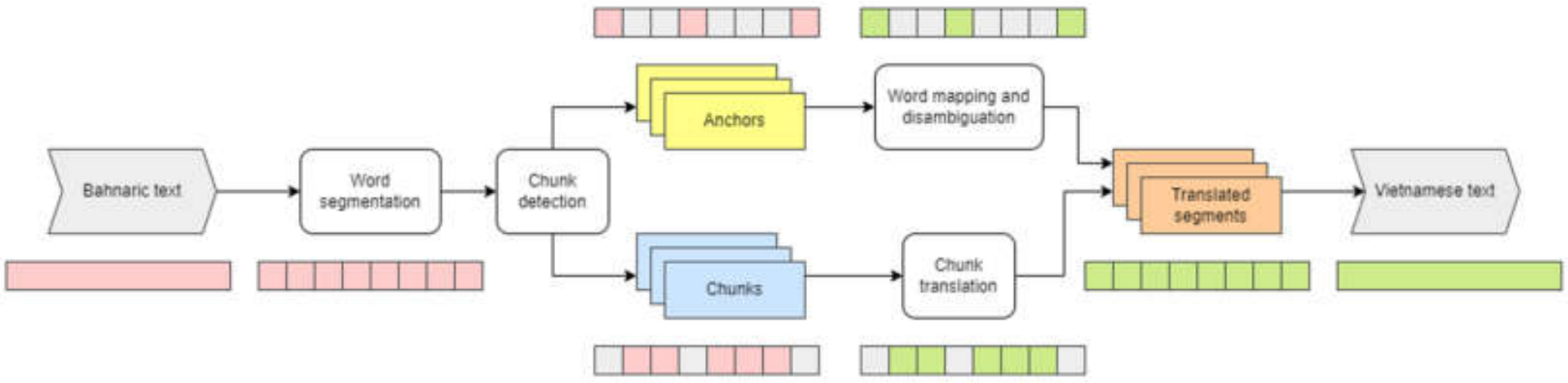

**Figure 1.** Overview of pipeline

## 2.2. Bahnaric-Vietnamese Fine-tuned BV-BARTpho

BART (Lewis et al. 2020) [9] is a sequence-to-sequence model which is capable of generating text from text, such as for tasks like summarization, translation, or text generation. By pre-training on a denoising autoencoder task, BART captures robust contextual information and effectively reconstructs corrupted text inputs, enhancing its understanding of language semantics. This pre-training includes a masked language model component, where parts of the text are masked and the model learns to predict the missing parts, further improving its language comprehension. The BARTpho model, an extension of BART tailored for the Vietnamese language, leverages the same bidirectional and autoregressive principles but is specifically pre-trained on large Vietnamese text corpora. By doing so, BARTpho captures the unique syntactic and semantic nuances of Vietnamese, which enhances its ability to understand and generate accurate translations. From that, authors developed BV-BARTpho by fine-tuning BARTpho with a Bahnaric language dataset, aiming to enhance performance on Bahnaric-Vietnamese translation tasks.

- **Dataset:** The dataset for fine-tuning process includes parallel corpora and a billingual dictionary. The parallel corpora are multiple pairs of sentences in both languages that are aligned at the sentence level. The billingual dictionary is a compilation of words and their translations from Bahnaric to Vietnamese language.
- **Training procedure:** To fine-tune the BARTpho model, we process a Bahnaric chunk identified from the segmentation phase as an input and encodes it into a latent representation. The decoder takes it and generates an equivalent Vietnamese chunk as an output. The model is trained to minimize the cross-entropy loss between the output and the target chunk.

## 2.3. Segmentation phase

In this phase, due to the lack of available resources for the Bahnaric language, we developed a dedicated word segmentation system specifically for this language.

- **Word Segmentation:** Similar to Vietnamese, a meaningful Bahnaric word is often composed of two or more individual words. Therefore, we utilized the parallel corpora of Bahnaric to identify the most frequent adjacent word groups and stored them in JSON format. Whenever the word segmentation is processed, the sentence will be prioritized to split into the most frequently occurring words from the output JSON file of the preprocessing. Figure 2 shows how our word segmentation be implemented.
- **Chunk Detection:** As introduced, chunks are Bahnaric words which cannot be directly mapped to Vietnamese due to the absence of its meaning in the billingual dictionary, and are also not punctuation marks or numbers. Thus, after identifing all anchors, the rest of the sentence will be chunks.

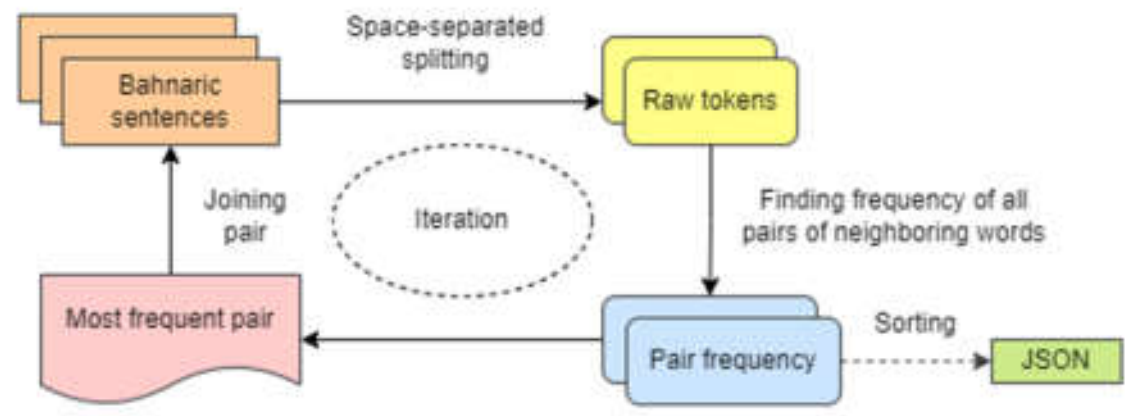

**Figure 2.** Word segmentation process

## 2.4. Mapping phase

In this phase, our model translates all Bahnaric anchors and chunks to Vietnamese segments. As the authors expected, there are cases where a Bahnaric word could be translated into multiple Vietnamese words from dictionary. To resolve this issue, we utilized the parallel corpus of Vietnamese to identify the neighboring words of each vocabulary item within a fixed-size window. Whenever a Bahnaric word has multiple meanings, we check if the neighboring words of the Vietnamese translation are present in the corresponding list of the candidate word we previously identified. If they are, each word is assigned a weight inversely proportional to its distance from the candidate word. The candidate Vietnamese word with the highest total score is then selected. Figure 3 indicates the overview of our method, which is called "Word disambiguation process".

For punctuation marks or numbers, they are all kept unchanged. The Bahnaric chunks are fed into the BV-BARTpho as an input and obtain the output as Vietnamese chunks.

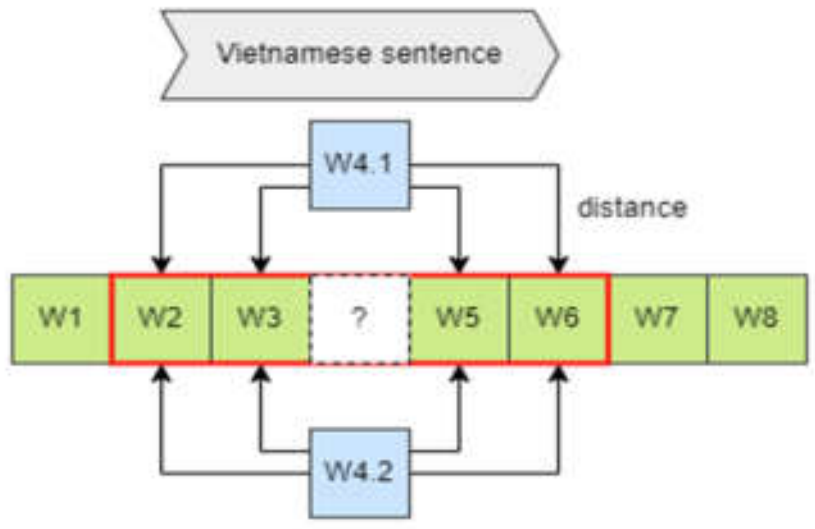

**Figure 3.** Word disambiguation process

## 2.5. Data Augmentation

**Multi-task Learning Data Augmentation (MTL DA)**: Inspired by two studies: EDA (Wei, Jason & Zou 2019) [10] and DA operations from a multi-task learning approach (Sanchez-Cartagena et al. 2021) [11]. This approach uses a set of simple DA transformations to generate synthetic target sentences, thereby enhancing the encoder. Transformations are controlled by a hyperparameter $\alpha$, which determines the percentage of target words affected by the transformation. Given $t$ as the number of tokens in our target sentences, brief explanations of each auxiliary task are shown below:

- **Swap:** Randomly swapping $\alpha \cdot t$ target words.
- **Token:** Randomly replacing $\alpha \cdot t$ target words with a special (UNK) token.
- **Source:** Copying the source sentence.
- **Reverse:** Reversing the word order of the target sentence.
- **Replace:** Randomly substituting $\alpha \cdot t$ target words and their corresponding aligned source words.

**Sentence Boundary Augmentation:** Originally inspired by Li et al. 2021 [12]. This method splits the target sentence into two parts. For two adjacent sentences, their parts are swapped and combined with the other parts. Luque 2019 pointed out that swapping at the sentence level can still maintain the semantic meaning [13]. Hence, sentence boundary augmentation has been applied in low-resource translation contexts, exposing the model to bad segmentation during training.

# 3. Results and Discussion

## 3.1. Dataset

Our research relied on extensive fieldwork to collect reliable data from Bahnar-speaking communities in Binh Dinh, Gia Lai, and Kon Tum. We focused on Binh Dinh due to the linguistic unity of its Bahnar dialects, which are easily understood by speakers in Gia Lai and Kon Tum. This central location provided deep access to Bahnar language culture and facilitated comprehensive data collection. The dataset includes Bahnar handwritten words from the Binh Dinh area (digitized from a bilingual dictionary), translated voice recordings from Vinh Thanh radio station, documents on various topics written in Bahnar, and a Bahnar dialect handbook.

We collected common sentences, formal and informal conversations, narrative stories, and folktales in Bahnar. The dataset is divided into a training set (16,105 sentence pairs), a test set (1,988 pairs), and a validation set (1,987 pairs).

## 3.2. Experiment Settings

### 3.2.1. Applying Data Augmentation

The training dataset was expanded by augmentation, resulting in a final training dataset that is twice the size of the original. As the augmented data is evaluated across multiple models, the augmentation process is performed iteratively, generating a new training dataset for each evaluation. This repetitive augmentation is designed to assess the proposed models' ability to adapt to variations within the dataset, providing insights into their robustness and adaptability to diverse linguistic instances in the low-resource Bahnar language.

In the augmentation process, we conducted with various hyperparameter $\alpha$ values spanning the range from 0.0 to 1.0. The authors observed that a value of 0.5 showcases its suitability for maintaining a balance between augmentation-induced diversity and dataset stability. Besides, our choice of $\alpha$ value at 0.5 serves as a deliberate choice to strike a balanced augmentation approach. With an alpha value of 0.5, the augmentation process introduces sufficient variability into the dataset, capturing diverse perspectives of the Bahnar language. Additionally, due to limited resources for both training and augmenting data, adopting a cautious and proactive approach, an alpha value of 0.5 is chosen as a safe and resource-efficient strategy.

The bilingual dictionary houses 13,029 words aligning both the Bahnar language and Vietnamese dictionaries. After augmenting the original training dataset, the expanded corpus undergoes a thoughtful division. This refined dataset undergoes strategic partitioning, allocating 80% for training with sentence chunking is activated and executed. The remaining 20% of the dataset is intended for training with complete sentences, ensuring a comprehensive approach that embraces both segmented and intact linguistic contexts.

### 3.2.2. Baselines

To assess the performance of our model, we conducted five experiments, wherein four baseline models were employed for comparison against our proposed model. The baselines include:

- **Transformer:** We reproduce a full stack of 6 encoder and decoder layers in the vanilla Transformer.
- **PhoBERT-fused NMT:** A BERT-fused NMT model with the replacement of PhoBERT.
- **PE-PD-PGN:** A Pretrained Encoder - Pretrained Decoder-fused (PE-PD) NMT model combined with the masked Pointer Generator Network.
- **BV-BARTpho**: Our proposed model without using Bahnaric tokenizer

### 3.2.3. Setting

In the training phase, we maintain a dropout rate of 0.1 and aggregate gradients over 64 batches, each comprising 4 instances, per weight update. To establish a sensible boundary, we capture the maximum length of both source and target sequences at 256, and introduce label smoothing with $\epsilon_{ls}$ set at 0.1. Our choice of optimization tools, including the Adam optimizer and learning rate scheduler, adheres to the conventions outlined by Zhu et al. (2020) [14]. All models are trained through 21 epochs, involving approximately 4,000 weight updates for each model.

### 3.2.4. Result

After conducting experiments, the authors computed the corresponding BLEU scores for each model, as shown in Table 1. It is evident from the results that the BV-BARTpho model with Bahnaric tokenizer achieved the highest BLEU score, indicating its efficiency in our considered methodology.

| Model | BLEU |
|---|---|
| **Transformer** | 20.51 |
| **PhoBERT-fused NMT** | 25.20 |
| **PE-PD-PGN** | 30.24 |
| **BV-BARTpho** | 28.44 |
| **BV-BARTpho (with Bahnaric tokenizer)** | **33.58** |

**Table 1.** BLEU Scores for each model

These results demonstrates a reasonable performance of the BV-BARTpho model in our research. Since Bahnar language is a low-resource language, it has a limited vocabulary and few synonyms. This means that when translating from a low-resource language like Bahnaric to Vietnamese, there are more variations in the possible translations for each *n*-gram. As a result, it is easier for our model to generate accurate translations and achieve the highest BLEU score comparing to the baselines.

Additionally, authors conducted experiments for the translation from Vietnamese to Bahnar to evaluate the effectiveness of applying each of the MTL DA auxiliary tasks, with the combination of the best-performing ones and the sentence boundary approach. Table 2 showcases the translation performance, measured in terms of BLEU score prediction.

| Method | BLEU |
|---|---|
| Baseline | 33.58 |
| **Swap** | **40.72** |
| **Token** | **39.83** |
| Source | 3.81 |
| Reverse | 23.75 |
| Replace | 44.01 |
| swap+replace | 48.23 |
| **swap+token** | **49.61** |
| replace+token+swap | 43.91 |
| **sentence boundary** | **50.72** |

**Table 2.** BLEU scores obtained with the baseline; MTL DA approach, using different auxiliary tasks and combinations of them, sentence boundary augmentation

To begin with, the baseline is the evaluation result when training and testing without any augmentation method applied. The result shows that the MTL DA approach consistently outperforms the baseline system except for the method “source”. In general, the auxiliary tasks “swap”, “token”, and “replace” are the best-performing ones. “Reverse” may give a lower performance result than these three methods above, suggesting that abnormal word order could negatively influence the main task. In contrast, “source” has the worst performance, indicating that the translation task could be affected by introducing a completely different vocabulary in the target.

Interestingly, using two of the three best auxiliary tasks together further improves performance, achieving higher BLEU scores of 48.23 (“swap”+”replace”) and 49.61 (“swap”+”token”) points than the baseline. The combination of “swap” and “token” methods gives the best performance. Although the authors has combined all three best methods, the results still cannot outperform the combination of token and swap. Overall, all combinations have improved the BLEU score implying that various auxiliary tasks impact the encoder in distinct manners and exhibit a sense of complementarity.

Considering the same training configuration, the result of sentence boundary augmentation method can be paired with any auxiliary task from MTL DA and their combinations. It indicates that sentence boundary augmentation method can perform well in the context of low-resource translation, and can also utilize the available limited resources. For low-resource machine translation augmentation, the nosing-based method applied at the phrase or sentence level has demonstrated superior performance compared to the nosing-based method applied at the word level.

## 4. Conclusions

Our approach has yielded good results for the Bahnaric - Vietnamese machine translation task, overcoming the resource scarcity barrier through multiple appropriate data augmentation strategies. This results supports the cultural linkage between the two ethnic groups, allowing it to expand and develop in the future.

***Acknowledgement*:** This research is funded by Office for International Study Programs (OISP), Ho Chi Minh City University of Technology (HCMUT), VNU-HCM under grant number **SVKSTN-2023-KH&KTMT-44**.

We acknowledge the support of time and facilities from HCMUT, VNU-HCM for this study.